\documentclass[letterpaper, 10pt, journal]{IEEEtran}
\IEEEoverridecommandlockouts
\usepackage[latin1]{inputenc}

\usepackage{paralist}
\usepackage{multirow}
\usepackage{cite}
\usepackage{amsmath,amssymb,amsfonts}
\usepackage{algorithm}
\usepackage{algorithmic}
\usepackage{graphicx}
\usepackage{textcomp}
\usepackage{float}

\usepackage{subfigure}
\usepackage{epstopdf}
\usepackage{cases}

\usepackage{verbatim}

\usepackage{float}

\usepackage{color,soul}%é«äº®å
\soulregister\cite7 % éå¯¹\citeå½ä»¤
\soulregister\citep7 % éå¯¹\citepå½ä»¤
\soulregister\citet7 % éå¯¹\citetå½ä»¤
\soulregister\ref7 % éå¯¹\refå½ä»¤
\soulregister\pageref7 % éå¯¹\pagerefå½ä»¤
\setulcolor{bule} %è®¾ç½®ä¸åçº¿çé¢è²ä¸ºè
\setstcolor{yellow} %è®¾ç½®overstrikingé¢è²ä¸ºé»
\sethlcolor{yellow} %è®¾ç½®é«äº®æ¾ç¤ºä¸ºç»¿
\usepackage{bm}
\def\BibTeX{{\rm B\kern-.05em{\sc i\kern-.025em b}\kern-.08em
    T\kern-.1667em\lower.7ex\hbox{E}\kern-.125emX}}
\begin{document}

\title{
Attention Loss Adjusted Prioritized Experience Replay}

\iffalse
\author{Zhuoying~Chen, Huiping~Li,~\IEEEmembership{Senior Member,~IEEE}, Rizhong~Wang
	%~\IEEEmembership{Life~Fellow,~IEEE}% <-this % stops a space
}

\fi

\author{
	\IEEEauthorblockN{Zhuoying~Chen$^{1}$, Huiping~Li$^{1}$~\IEEEmembership{Senior Member,~IEEE}, Rizhong~Wang$^1$}
	
	\IEEEauthorblockA{$^1$ School of Marine Science and Technology,
		Northwestern Polytechnical University,  Xi'an, China, 710072}
	
	\IEEEauthorblockA{Email:czysmile@mail.nwpu.edu.cn, lihuiping@nwpu.edu.cn, rizhongwang@mail.nwpu.edu.cn}
}

\maketitle
\begin{abstract}
Prioritized Experience Replay (PER) is a technical means of deep reinforcement learning by selecting experience samples with more knowledge quantity to improve the training rate of neural network. However, the non-uniform sampling used in PER inevitably shifts the state-action space distribution and brings the estimation error of Q-value function. In this paper, an Attention Loss Adjusted Prioritized (ALAP) Experience Replay algorithm is proposed, which integrates the improved Self-Attention network with Double-Sampling mechanism to fit the hyperparameter that can regulate the importance sampling weights to eliminate the estimation error caused by PER. In order to verify the effectiveness and generality of the algorithm, the ALAP is tested with value-function based, policy-gradient based and multi-agent reinforcement learning algorithms in OPENAI gym, and comparison studies verify the advantage and efficiency of the proposed training framework.

\iffalse
we combine LA2P with Deep Q learning (DQN), Deep Deterministic Policy Gradient (DDPG) and Multi-Agent Deep Deterministic Policy Gradient (MADDPG), respectively;

\fi

\end{abstract}


\section{Introduction}

In recent years, deep reinforcement learning has made great achievements in many fields such as go\cite{Silver2016}, natural language processing \cite{Xiao2018}, robot control \cite{Fan2020}, etc. Experience Replay mechanism \cite{Lin1992} is an important training means of deep reinforcement learning. It learns the neural network by randomly selecting a fixed number of experience samples from buffer, and breaks the relevance between training samples in  reinforcement learning. Experience Replay puts the samples in the experience pool with the same sampled frequency, and this uniform sampling method is bound to cause too much invalid exploration in the early stage of training. Moreover, when the agent is in the sparse reward environment \cite{Bing2023}, this disadvantage is rather obvious, which makes the algorithm difficult to converge.

In order to break the bottleneck of the above algorithm, the Priority Experience Replay (PER) mechanism  \cite{Schaul2016} is proposed; its core idea is to take the Temporal Difference (TD) error's absolute value as the index to measure the importance of the sample. By changing the probability distribution of the collected samples and learning the relatively important experience samples more frequently, the training speed and efficiency of the network have been improved. It is noted that the non-uniform sampling of PER artificially adjusts the sampled frequency of the samples, and shifts the state-action space distribution to estimate the Q-value function \cite{Mnih2013,Mnih2015,Sharma2021,Tesauro2004} at the same time \cite{Yang2022}, which leads to the estimation error. Aiming at error correction, a hyperparameter $\beta$ \cite{Schaul2016} \cite{Sovrano2022} is introduced to regulate the weight of importance sampling, which can prevent the instability of the training process, but it only keeps unbiased in the converged phase. Note that $\beta$ is positively related with training progress \cite{Schaul2016}, and it increases linearly to 1 from initial value $\beta_0$ as the number of training episodes increases. However, the training progress is not distributed uniformly over training episodes, and a linear change of $\beta$ would cause extra error. For example, when the episodes reach 50$\%$ of the total number, the model may have already converged. In this case, $\beta$ is linearly set to be $\frac{1-\beta_0}{2}$, much less than the desired value 1 for the converged model, which would introduce large estimation errors of Q-value function during training.

To resolve this issue, in this paper, an efficient and general training framework called Attention Loss Adjusted Prioritized (ALAP) Experience Replay is proposed. ALAP can greatly reduce the estimation error of the Q-value function from both active and passive aspects. On the one hand, the dynamic loss function is constructed by Huber equation \cite{Kaan2021,Fujimoto2020,Saglam2022}, which has passive adaptability in the face of different sizes of TD error samples. On the other hand, we introduce the Self-Attention mechanism to add parallel attention network branches into the actor-critic (A-C) framework or Q network. The attention module can actively measure the similarity of sample distribution from experience pool, which represents the exact progress of the training process. In order to establish the mapping relationship between $\beta$ and the output of attention module, we normalize the output to obtain more accurate $\beta$, so that the estimation error of Q-valued function can be actively reduced.

\iffalse
 Since there is a positive correlation between $\beta$ and training progress, the most important step of LA2P is to quantify the exact training progress in real time to fit a more accurate $\beta$. The agent primarily generates behavior through random exploration in early stages of training so that the sample distribution of experience pool is random and uniform. As the training proceeds, the exploratory behavior decreases, and the agent makes more use of the current optimal strategy to select actions with high reward. As a result, the similarity of the sample distribution will grow up as well. The attention module utilizes the projection principle between vectors to calculate the similarity of samples in experience pool, which represents the exact progress of the training process. In order to establish the mapping relationship between $\beta$ and the output of attention module, we normalize the output to obtain more accurate $\beta$, which can actively conpensate the estimation error in conventional PER.
\fi

Furthermore, considering that PER applys the fixed criteria to screen out experience samples, which cannot restore the original sample distribution from experience pool, we propose a Double-Sampling mechanism which covers both priority based sampling (PS) and random uniform sampling (RUS). PS is only in charge of providing training samples for critic network (or Q network), while RUS is responsible for offering data input of attention module.

To evaluate the algorithm's efficiency and generality, we compare ALAP with conventional PER and the state of art algorithm LAP in DQN \cite{Mnih2015}, DDPG \cite{Timothy2015} and MADDPG \cite{Lowe2017}, and the performance advantages are verified in three different environments of OPENAI gym\cite{Brockman2016}.

\section{Related Work}

The inspiration of priority sampling used in reinforcement learning comes from the prioritized sweeping for value function iterations, and its success in DQN \cite{Moore1993,David1997,Seijen2013} has attracted a lot of attention. DQN uses the same maximization operator to select and evaluate actions, which brings the problem of overestimation. The PER algorithm in the Double-DQN (DDQN) is developed in \cite{Tao2020}. In addition, PER has generated favorable outcomes in many algorithms such as DDPG \cite{Hou2017} \cite{Lu2022}, Rainbow \cite{Hessel2018}, etc.

PER itself has also evolved many improved versions. Liu \cite{Liu2018} and Vanseijen \cite{Vanseijen2018} studied the effect of buffer size on algorithm performance. In order to alleviate the waste of computing resources caused by the excessive experience pool, Shen \textit{et al}. \cite{Shen2019} proposed an experience classification method, which divides the TD error distribution into different segments, and the transition tuples in the same segment have the same priority. This clustering method can reduce the cache space, and break the relevance of experience samples. Aiming to screen out more valuable experience samples, Gao \cite{Gao2021} \textit{et al}. integrated the reward value with the TD error to form a new priority parameter, which replaces the TD error as screening basis.

The aforementioned works mainly focus on the adjustment of sample priority evaluation, while ignore the deviation caused by PER itself. In order to resolve this problem, a Loss Adjusted Prioritized (LAP) \cite{Fujimoto2020} algorithm is proposed, which proves that any loss function evaluated by non-uniform sampling can be transformed into another uniform sampling one with the same expected gradient. The loss function is segmentaly described by Huber function, and different sampling methods are adopted according to the variation of TD error. This can suppress the sensitivity of the root Mean Square Error (MSE) loss to outliers, and reduce the influence of Mean Absolute Error (MAE) loss on convergence rate. On the basis of LAP algorithm, the work \cite{Saglam2022} theoretically analyzes the reasons for the poor effect on the combination of PER and actor-critic algorithm in continuous action control environment, and claims that there is a big error between Q-function calculation and the actual value, so it cannot correctly guide the actor network to make reasonable actions. 

The LAP algorithm improves the robustness aganist outliers by designing the segemental loss function. However, the non-uniform sampling of PER will change the state-action space of the model when the MAE loss is applied, resulting in the estimation deviation of Q-value function. Therefore, LAP does not solve the underlying problem of PER.

\iffalse
In the previous research, PER technology is mostly used in single-agent algorithms such as DQN, DDQN, DDPG, etc. Considering that multi-agent algorithm has a wider application scene, the combination of PER and multi-agent algorithm obviously has more practical worth. With the increase of agent numbers, the input dimension of Q-value function grows exponentially, the disaster of dimensionality will inevitably lead to exceeding the limit for computing load that makes the algorithm non-convergent. Iqbal et al. puts forward a Multi-Actor-Attention-Critic (MAAC) \cite{Iqbal2019} algorithm, which uses the attention mechanism \cite{Vaswani2017} to select important information from the observation-action pairs of other agents to construct the critic input of each agent, which increases the scalability of the algorithm. Based on the idea of MAAC, this paper considers from another perspective on how to use the characteristics of the similarity calculation in attention mechanism to correct the deviation caused by PER when the state dimension is fixed, so as to improve the training efficiency of the algorithm.

\fi

In this paper, we propose a new method to resolve this issue by accurately determining the relation between the training progress and $\beta$. In particular, this paper first designs an improved Self-Attention network, which quantifies the training progress by calculating the similarity of samples in experience pool. Meanwhile, we propose a Double-Sampling mechanism to ensure the parallel execution of training and similarity calculation. In this way, the deviation of the Q-value estimation can be greatly reduced. The main innovation of this study is as follows:

\begin{itemize}

\item An extended actor-critic framework (or DQN) is proposed, which embeds the attention module into the A-C strcture (or DQN) as a parallel branch of the critic network (or Q network), takes the sequence of state-action pairs collected from experience pool as the input of attention module to calculates the similarity of the experience samples, and constructs the nonlinear mapping relationship between the output of the attention module and $\beta$ through the fully connected network.

\item A parallel Double-Sampling Mechanism (DSM) for two sampling procedures is proposed. The priority based sampling (PS) provides training samples for the model, and the random uniform sampling (RUS) is responsible for providing the data input of attention module. In addition, in order to guarantee the stable parallel execution of the Double-Sampling mechanism, we build a mirror buffer with the same data distribution as the original experience pool for the data source of uniform sampling.

\item In addition, an imporved Huber loss function is proposed to describe the loss function in segments, which extremly suppress the model sensitivity to outliers by adaptive loss; meanwhile, the priority clipping part of the LAP is removed to ensure the algorithm will not sacrifice the training speed.

\end{itemize}

The structure of the rest paper is as follows: Section \ref{PRELIMINARIES} is the problem statement and the introduction of the basic knowledge, and Section \ref{section:LA2P} describes the design process of ALAP algorithm in detail. Section \ref{section:experiment} provides comparative experimental results and analysis. Finally, Section \ref{conclusion} draws the conclusions.

\begin{comment}   %%%%Notations??
Notations: Let $\mathbb R$ and $\mathbb{N}_{\geqslant0}$ denote the real number and nonnegative integers, respectively.  $\mathbb R^n$ is the set of all real $n$-dimensional column vectors. For a matrix $A$, $\bar{\lambda}(A)$ and  $\underline{\lambda}(A)$ represent maximum and minimum eigenvalues of $A$, respectively.
For two nonempty sets $p$ and $q$, $p\wedge q$ denotes the conjunction of $p$ and $q$.
$diag(x_1,x_2,\cdots x_n)$ indicates a diagonal matrix with elements $x_1,x_2,\cdots x_n$. For a vector $\bm{x}\in\mathbb R^n $ and a matrix $P$, we use the $P$-weighted norm ${\left\|\bm x\right\|}_P$ to denote $\sqrt{{\bm x}^T P \bm x}$ and $P^{-1}$ as the inverse operation of $P$. We call the function $y(\bm x)$ is locally Lipschitz continuous in $\bm x\in \Gamma\subseteq \mathbb{R}^n $, if the inequality $\left\|y(\bm x_1)-y(\bm x_2)\right\|\leqslant L_y\left\|\bm x_1-\bm x_2\right\|$ is satisfied with a positive constant $L_y$ for all $ \bm x_1,\bm x_2 \in \Gamma$.
\end{comment}

\section{PRELIMINARIES}\label{PRELIMINARIES}

\subsection{Markov Decision Process}\label{subsection:System Description}

A Markov model with five key elements $\langle S,A,S',r,\gamma \rangle$ is defined to describe the interaction process between the agent and environment. Among them, the state $S$ represents the possibility configuration of the agent, while vector $A$ denotes the action space. At each time step $t$, the agent follows the strategy $\pi_{\theta}$ to select the action $A_t$, and obtains the next state $S'$ according to the state transition equation. In this process, the agent received immediate reward $r_{t}$ according to a function of its action and state. We call the process of environment from the initial state to the terminated state as a trajectory $\zeta$, and the agent's goal is to maximize the expected return $R=\sum_{t=0}^{T}\gamma_tr_{t}$ to continuously optimize $\zeta$, where $\gamma$ and $T$ represent the discount factor and time horizon, respectively.

\subsection{Prioritized Experience Replay}\label{PER}
PER applys a non-uniform sampling strategy, which adjusts the sampled frequency of experience transitions depending on the magnitude of TD-error $\left|\delta\right|$.
It has two improvements in comparison with the uniform sampling experience replay \cite{Kang2021}. Firstly, PER assigns each experience sample $i$ with a sampling probability proportional to its TD error, and improves the training efficiency by learning important experience samples more frequently as follow:

\iffalse
\begin{equation}\label{prioritiy_distribution_function}
	\begin{aligned}
		P(i)=\frac{\left|\delta(i)\right|^{\alpha}+\epsilon}{\sum_{j}(\left|\delta(i)\right|^{\alpha}+\epsilon)}
	\end{aligned}
\end{equation}
\fi

\begin{equation}\label{prioritiy_distribution_function}
	\begin{aligned}
		P(i)=\frac{p_i^{\alpha}}{\sum_{j}p_j^{\alpha}}
	\end{aligned}
\end{equation}

In the formular (1), $p_i=\left|\delta(i)\right|+\epsilon$ is the priority of transition $i$. The hyperparameter $\alpha$ is used to smooth out extremes, and a minimal positive constant value $\epsilon$ makes sure that the sampled probability is greater than 0, which gives all transitions a chance to be selected.

Secondly, it introduces importance sampling weight ratios $\bar{w}(i)$ to correct the estimation deviation caused by shifted state-action distribution:

\begin{equation}\label{importance sample weights}
	\begin{aligned}
		\bar{w}(i)=\frac{w(i)}{\max_{j}w(j)},		w(i) = \left( \frac{1}{N}\cdot\frac{1}{P(i)} \right)^\beta,
	\end{aligned}
\end{equation}

\begin{equation}\label{importance sample weights}
	\begin{aligned}
			L_{PER}(\delta(i))=\bar{w}(i)L(\delta(i)).
	\end{aligned}
\end{equation}

Note that $\bar{w}(i)\delta(i)$ will be used instead of $\delta(i)$ when updating formular (3). The hyperparameter $\beta$ increases linearly from $\beta_0$ to 1 along the training process.

\subsection{Loss Adjust Prioritized Experienced Replay}\label{LAP}

The linear annealing of importance weight in PER cannot completely eliminate the deviation, and the  sensitivity to outliers is easy to further magnify the deviation. Fujimoto \textit{et al}. \cite{Fujimoto2020} proposed an LAP algorithm, which uses the segemental loss function 
\begin{equation}
	L_{Huber}(\delta(i))=
\begin{cases}
	0.5\delta(i)^2 & 	\left|\delta(i)\right|\leq 1,\\
	\left|\delta(i)\right| & otherwise,\\
\end{cases}
\end{equation}
combined with priority clipping scheme to further reduce the deviation:
\begin{equation}\label{clipping}
	\begin{aligned}
		P(i)=\frac{max(\left|\delta(i)\right|^\alpha,1)}{\sum_{j}max(\left|\delta(j)\right|^\alpha,1)}.
	\end{aligned}
\end{equation}

By combining (4) and (5), it can be seen that when the absolute value of TD error is less than and equal to 1, MSE is applied as the loss function according to (4), and the priority of transition $i$ is cut up to 1, so that the uniform sampling is performed for each transition $i$ with a sampled probability of $\frac{1}{N}$ according to (5). Similarly, if the TD error is greater than 1, MAE is used to suppress the sensitivity to outliers in (4), and the non-uniform sampling is performed according to (5). Although LAP avoids the interference of outliers on training, the uniform sampling part decreases the training speed. In addition, MAE can only suppress the sensitivity to outliers, but cannot correct the shifted distribution, and the deviation does exist. Saglam \textit{et al}. \cite{Saglam2022} proposed an LA3P algorithm based on LAP, using inverse sampling to select samples with small TD error to train the network, which avoids the uncertainty caused by large TD error transition, but does not meet the requirement on sample knowledge of PER.

\section{Attention Loss Adjusted Prioritized Experience Replay}\label{section:LA2P}

\iffalse
Different from the existing algorithms, our proposed LA2P algorithm will solve the deviation in Q-value function esitimation caused by PER from the root without sacrificing the training speed. We employ Self-Attention network combined with Double-Sampling mechanism to calculate the similarity of sample distribution in experience pool to quantify the training progress, so as to get the real time accurate value of $\beta$ to dynamically correct the deviation.
\fi

In this section, the design of ALAP algorithm is described in detail from two parts: Self-Attention network and Double-Sampling mechanism.

\subsection{Design of Self-Attention Network}\label{self-attention}

In PER and its improved algorithms, the hyperparameter $\beta$ is an important index to regulate the importance sampling weight (IS), which determines the correction strength of the algorithm with respect to error caused by PER. $\beta$ is positively correlated with the training progress and reaches 1 to fully compensate for esitmation error as the training ends. The PER uses the linear annealing method to make $\beta$ increase linearly with the training episodes. However, the training progress is not uniformly distributed over the scale of episodes number, and the way of linear parameter tuning may exacerbate the error. Note that the size of $\beta$ depends on the specific progress of the training; in order to quantify the training progress, we build an improved Self-Attention network to measure the training progress by calculating the similarity of sample distribution in the experience pool. At the beginning of training, the transitions in experience pool generated by the stochastic exploration of agent is random and uniform. As the training proceeds, the agent will make more use of the current optimal strategy to select high-return actions, that is to say, given a state, the action selected by the agent will be roughly the same, and the similarity of transitions sequence generated by the interaction between agent and environment will be improved. We input the mini-batch sized state-action pair sequences $X=[(S_1,A_1),(S_2,A_2),...,(S_m,A_m)]$ into the Self-Attention network, whose output is the quantized value of the sample similarity in experience pool, and $\beta$ is obtained after normalization. 

The attention function can usually be described as a nonlinear fitting network that maps a query and a set of key-value pairs to the output \cite{Iqbal2019,Vaswani2017}. Its output is essentially a weighted sum containing importance weights. In practice, we package queries, keys and values into the corresponding matrices Q, K and V, and the output of attention value is calculated as follows:

\begin{equation}\label{attention}
	\begin{aligned}
		Attention(Q,K,V)=softmax(\frac{Q \cdot K^T}{\sqrt{d_k}})V,
	\end{aligned}
\end{equation}
where $d_k$ is the dimension of queries and keys vectors, dividing the dot-product $Q \cdot K^T$ by $\sqrt{d_k}$ to prevent the gradient from disappearing.

\begin{figure}[htbp] %b
	\begin{minipage}{1\linewidth}
		\centering
		\includegraphics[width=0.8\textwidth]{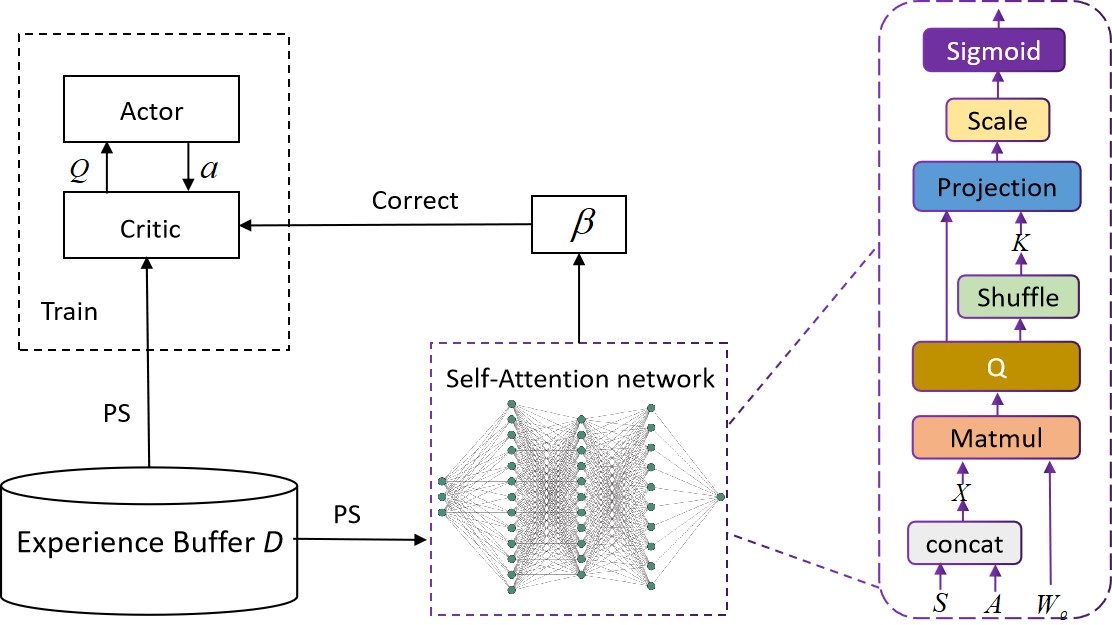}
	\end{minipage}
	\caption{Improved Self-Attemtion network.}
	\label{Self_Attention}
\end{figure}

In order to meet the needs of similarity calculation on internal elements of the input sequence, this paper optimizes the Self-Attention network properly and removes the value vector, and the attention module only outputs the similarity of key-query pairs. Considering that X is a list constructed of m state-action pair vectors, referring to the physical meaning of vector projection, we measure the similarity between $Q$ and $K$ by the sum of the projections among the corresponding vectors, where $Q=XW_Q$, $K=shuffle(Q)$, and both of them maintain dimensions consistent with X; $W_Q$ represents the initial weight matrix of $Q$. Since the similarity is calculated for the internal elements of  $X$, we need to do projection operation after randomly rearranging the order of the elements (by shuffle operation) in $Q$ for each $q_i$ and $k_i$. The process for improved Self-Attention network to calculate the attention value is shown as follows:

\begin{equation}\label{self-attention}
	\begin{aligned}
		Self-Attention(Q,K)=sigmoid(\frac{Q \bullet K^T}{\sqrt{d_k}}),
	\end{aligned}
\end{equation}

\begin{equation}\label{self-attention}
	\begin{aligned}
		Q \bullet K^T= \sum_{i=1}^{m}\frac{(q_i\cdot k_i)}{\left|k_i\right|}
		,
	\end{aligned}
\end{equation}
where $\bullet$ stands for projection sum operation of the corresponding elements between $Q$ and $K$. The input of the Self-Attention module is the state-action pair sequence, and the output attention value represents the similarity of the elements in $X$. After passing through the fully connection layer, the attention value is normalized by the activation function to obtain $\beta$. The schematic diagram of Self-Attention mechanism is shown in Fig. \ref{Self_Attention}. Note that the schematic diagram in this paper only depicts the network structure of ALAP under A-C framework, and the combination of ALAP with the algorithm based on value-function is similar and it is omitted here.

We can see that the Attention network's input is a mini-batch sized sequence of experience transitions provided by PS in Fig.\ref{Self_Attention}. However, PS exploits the fixed standard to screen out samples, thus its sample distribution cannot restore the real situation in experience pool, which inspires us to design the Double-Sampling mechanism.

\subsection{Design of Double-Sampling Mechanism}\label{LAP}

The core idea of ALAP is to quantitatively describe the training progress by calculating the similarity of the sample distribution in experience pool through the attention module. Therefore, the sample distribution from data source of the attention module should be consistent with the sample distribution in experience pool. However, the priority based sampling way artificially changes sample's visited frequency, so that the data source obtained by PS cannot reflect the real situation of data distribution in whole experience pool. Hence, we need to rely on random uniform sampling to provide data input for the attention module. In addition, the model network still needs PS to provide high-quality training samples. To sum up, we propose a parallel sampling method called Double-Sampling mechanism, which covers both PS and RUS. For ease of description, we define the data source acquired by PS as $s$ and the data source obtained by RUS as $s^{*}$. 

In the training process, the priority based sampling provides data samples for training model, and the uniform sampling is only responsible for providing data as input for attention module to fit $\beta$, which can correct the estimation error in real time. The key point of the Double-Sampling mechanism is not to interfere with the normal operation of training, and to collect data in the same experience pool $D$ using both sampling procedures simultaneously. Since the two sampling procedures are done simultaneously, even though the sampled batchsize is much smaller than the volume of the experience pool $D$, there is still a certain probability that the same sample will be selected for both sampling methods, resulting in instability of the algorithm. As a result, we propose a concept of mirror buffer $D^*$ and construct a mirror experience pool with the same data distribution as the original one. At each iteration step, the mirror buffer $D^*$ adds and updates experience samples at the same time as the original experience pool $D$ to keep the data synchronized.

\begin{figure}[htbp] %b
 	\begin{minipage}{1\linewidth}
 		\centering
 		\includegraphics[width=0.8\textwidth]{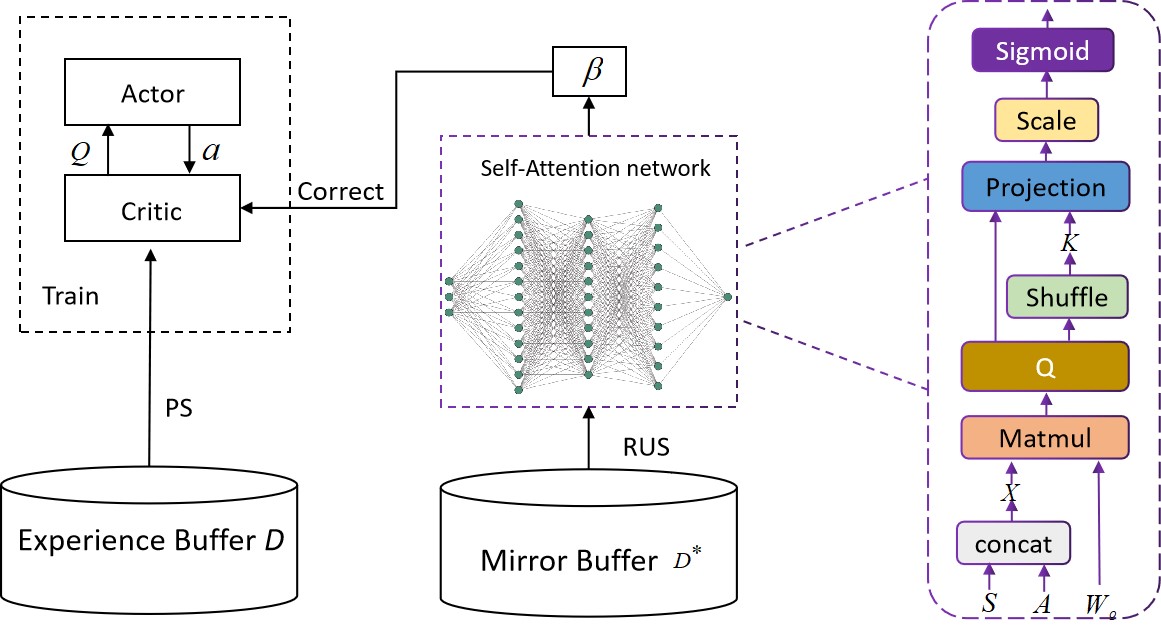}
 	\end{minipage}
 	\caption{Double-Sampling mechanism.}
 	\label{Double}
 \end{figure}

It is worth noting that, although from the physical level, the two sampling steps are carried out in two different buffers, but from the data level, they are actually operated in the same experience pool. The schematic diagram of Double-Sampling mechanism is shown in Fig.\ref{Double}.

\begin{algorithm}[htbp] %h
	\caption{ALAP Algorithm}
	\begin{algorithmic}[1]
		\renewcommand{\algorithmicrequire}{\textbf{Input:}}
		\renewcommand{\algorithmicensure}{\textbf{Output:}}
		\REQUIRE mini-batch  $m$, step-size $\sigma$, replay period $K$, buffer sotrage $N$, exponent $\alpha$ and $\beta$, budget $T$, tiny positive contant $\epsilon$.
		\STATE Initialize replay buffer $D = \emptyset$, mirror replay buffer $D^* = \emptyset$, $p_1=1$, $\Delta=0$
		\STATE Observe environment state $S_0$ and choose action $A_0$
		\FOR{t=1 \TO $T$}
		\STATE Observe $S_t$, $r_t$, $\gamma_t$
		\STATE Store transition $(S_{t-1},A_{t-1},r_t,\gamma_t,S_t)$ in $D$
		\STATE Store transition $(S_{t-1},A_{t-1},r_t,\gamma_t,S_t)$ in $D^*$
		\IF {$t > K$}
		\FOR {t=1 \TO $K$}
		\STATE Sample transition $j \sim P(j)=p_j^\alpha/\sum_{i}p_i^\alpha$
		\STATE Compute TD-error $\delta_j$ and update transition priority $p_j\gets \left|\delta(j)\right| + \epsilon$
		\STATE Obtain transitions $s$ with mini-batch $m$ from $D$ by PS for model training
		\STATE Obtain transitions $s^{*}$ with mini-batch $m$ from $D^*$ by RUS 
		\STATE Compute $\beta$ through Self-Attention network according to $s^*$
		\STATE Compute importance sampling weight $\bar{w}(i)=(N\cdot P(j))^{-\beta}/\max_{i}w(i)$
		\STATE Compute weight-change $\Delta \gets \Delta +\bar{w}(j)\cdot \delta(j) \cdot \nabla_{\theta}Q(S_{t-1},A_{t-1})$
		\ENDFOR
		\STATE Update weights $\theta \gets \theta+\sigma \cdot \Delta$, reset $\Delta=0$
		\STATE Copy parametes to target network $\theta_{target} \gets \theta$
		\ENDIF
		\STATE Choose action $A(t) \sim \pi_{\theta}(S_t)$
		\ENDFOR

	\end{algorithmic}
\end{algorithm}

Finally, in order to reduce the sensitivity of the algorithm to outliers, ALAP follows the Huber loss function applied in LAP and LA3P; unlike them, we remove the sample priority clipping part and use PER-sampled data to train the model throughout the whole process to ensure unbiased Q-value function estimation without sacrificing training speed. The overall operation flow of ALAP is shown in \textbf{Algorithm 1}.

\section{Experiment}\label{section:experiment}

In order to verify the effectiveness and generality, we integrate ALAP with DQN and DDPG, and test them in the environments \textbf{cartpole-v0} \cite{Kanno2022} and \textbf{simple} \cite{Lowe2017} from \textbf{OPAIgym}, respectively. To further demonstrate the versatility of the algorithm, we combine ALAP with multi-agent algorithm MADDPG, and carry out several groups of comparative experiments in \textbf{simple$\_$tag} \cite{Lowe2017}, and the typical multi-agent confrontation environment of \textbf{MPE}. In each environment, we use the same network structure, reward shaping and hyperparameter configuration to train the algorithms for comparison. \iffalse The primary distinction between the improved algorithm and the baseline is mainly reflected in the different sampling ways and error compensation strategies.\fi Additionally, we test the algorithms under different mini-batch sizes in each environment to compare their performance.

\subsection{Results with DQN}\label{DQN}

The environment \textbf{cartpole-v0} is utilized to test all the algorithms. In the comparison experiments, all algorithms execute different sampling manners to draw mini-batch samples from the corresponding experience pool of volume $2\times10^4$ for training the model, which is consturcted by 3-layer ReLU MLP with 24 units per layer. During the training, the agent chooses the action based on the $\epsilon-greedy$ strategy, the decay rate of $\epsilon$ is 0.0002, and a non-zero lower bound $\epsilon \geq$ 0.0001 is set to maintain the exploration ability of the agent. The total training episodes and the maximum simulation steps of each episode are both 200. The Adam optimizer is used to optimize the network parameters with a learning rate of 0.001 and a discount factor of 0.99. In \textbf{cartpole-v0}, the cartpole keeps the inverted pendulum upright through the left and right displacement, and any behavior of the agent, including the termination action, can be rewarded with a return value of 1. Since the maximum step per episode is 200, the maximum bonus the cart can get is 200 as well.

To avoid unintentional outcomes, we investigate the effectiveness of ALAP against other algorithms under batchsize=32, 64, 128, and we ran each case 20 times.

\iffalse
\begin{figure*}[htbp] %??htbp
	%\rotatebox{90}{\scriptsize{~~~~~~~~~~~~~~~~~~~~~~~reward}}
	\begin{minipage}[t]{0.33\linewidth} %??????????
		\centering
		\includegraphics[width=2.5in, height=1.9in]{NEWPLOT/car32.jpeg} %????????
		%\caption{Mean reward of value-function based algorithm} %?????
		\label{DDPG} %????
	\end{minipage}%
	\begin{minipage}[t]{0.33\linewidth}
		\centering
		\includegraphics[width=2.5in, height=1.9in]{NEWPLOT/car64.jpeg}
		%\caption{Simple$\_$tag}
		\label{fig2}
	\end{minipage}%
	\begin{minipage}[t]{0.33\linewidth}
		\centering
		\includegraphics[width=2.5in, height=1.9in]{NEWPLOT/car128.jpeg}
		%\caption{simple$\_$adversary}
		\label{fig3}
	\end{minipage}
	\caption{Mean reward of Policy gradient based algorithm} 
\end{figure*}

\fi

\begin{figure*}[!t]
	\centering
	\subfigure[batchsize=32]{
		\includegraphics[scale=0.34]{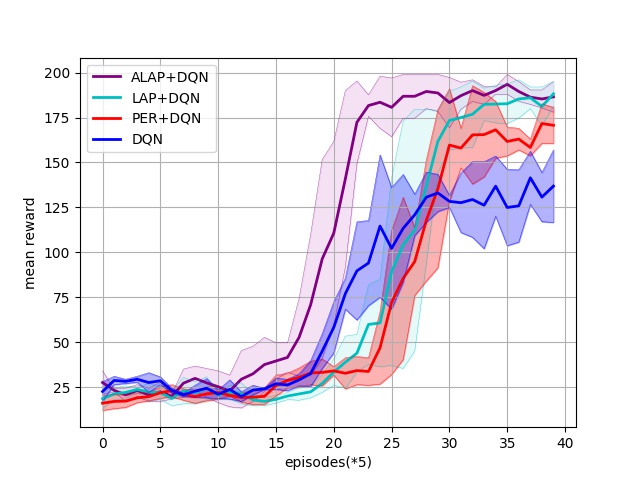}}
	\subfigure[batchsize=64]{
		\includegraphics[scale=0.34]{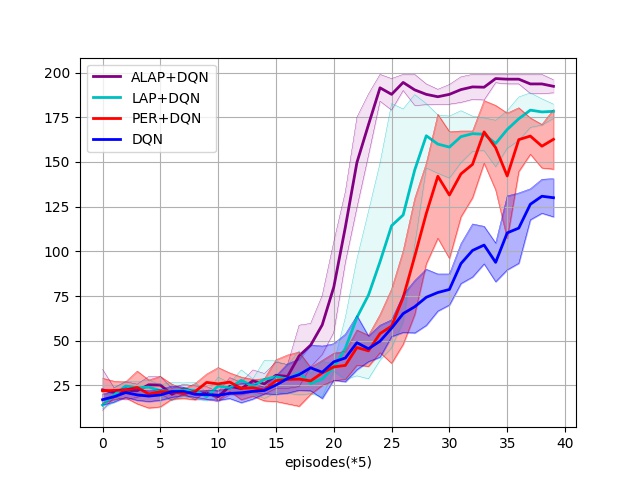}}
	\subfigure[batchsize=128]{
		\includegraphics[scale=0.34]{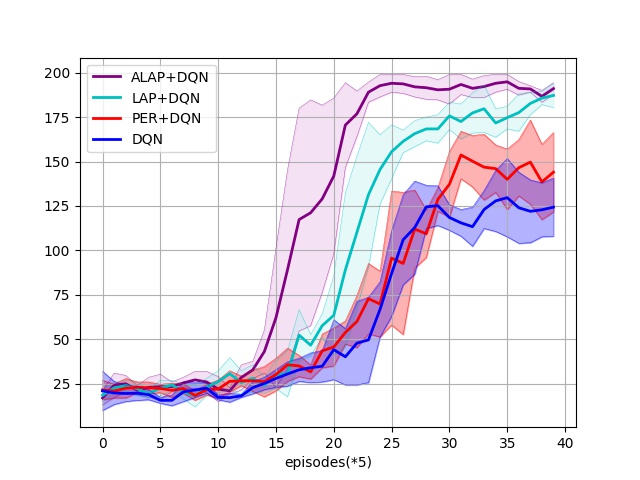}}
	\caption{Mean reward of value-function based algorithm.}
	\label{fig_3}
\end{figure*}

Fig.3 shows the average reward comparison curves between ALAP + DQN, LAP + DQN, PER + DQN and DQN. To compare the algorithm variance and stability, we set a 50$\%$ confidence interval, and drew the confidence bands of each curve, which is also applied in other two environments. It is clearly that the ALAP algorithm outperforms others in terms of training speed and average reward. The ALAP algorithm achieves steady state much faster and has a greater reward value. In terms of stability, the ALAP algorithm has a narrower confidence band during the whole training phase, indicating that it can efficiently minimize variation and is more resilient to outliers. The LAP algorithm is slightly inferior to ALAP in terms of convergence speed and quality because LAP partially utilizes uniformly sampled data to train the model at the expense of training speed, but it still provides some suppression to outliers. PER does have an improvement in convergence speed and reward peaks compared to DQN, but it has a large variance in the training process. \iffalse which constrains the training effect. \fi

\iffalse
In terms of robustness, LA2P still has an excellent performance and always outperforms the rest of the algorithms under different hyperparameter configurations; whereas the training speed of LAP and PER are even less fast than DQN at batchsize=32.
\fi

\subsection{Results with DDPG}\label{DDPG}

The scene \textbf{simple} is applied to evaluate the algorithms. Both original and mirror buffers are set with volume of $10^6$ to provide transitions for different sampling methods. The network model contains 2-layer ReLU MLP, with 64 neurons in each hidden layer. The total number of training episodes is 2000, the maximum simulation step is 25, the learning rate is 0.001, and the discount factor is 0.95. In this environment, there is just one agent and one obstacle, and the agent's reward depends inversely on how far away from the obstruction it is. We still trained 20 times under each batchsize to collect the comparison results.

\iffalse
 An experience pool with a capacity of $10^6$ is built to provide training samples for network model, which contains 2-layer ReLU MLP, with 64 neurons in each hidden layer. 
\fi

\iffalse
\begin{figure*}[htbp] %??htbp
	%\rotatebox{90}{\scriptsize{~~~~~~~~~~~~~~~~~~~~~~~reward}}
	\begin{minipage}[t]{0.33\linewidth} %??????????
		\centering
		\includegraphics[width=2.5in, height=1.9in]{NEWPLOT/simple32.jpeg} %????????
		%\caption{Mean reward of Policy gradient based algorithm} %?????
		\label{DDPG} %????
	\end{minipage}%
	\begin{minipage}[t]{0.33\linewidth}
		\centering
		\includegraphics[width=2.5in, height=1.9in]{NEWPLOT/simple64.jpeg}
		%\caption{Simple$\_$tag}
		\label{fig2}
	\end{minipage}%
	\begin{minipage}[t]{0.33\linewidth}
		\centering
		\includegraphics[width=2.5in, height=1.9in]{NEWPLOT/simple128.jpeg}
		%\caption{simple$\_$adversary}
		\label{fig3}
	\end{minipage}
\caption{Mean reward of Policy gradient based algorithm} 
\end{figure*}
\fi

\begin{figure*}[!t]
	\centering
	\subfigure[batchsize=32]{
		\includegraphics[scale=0.34]{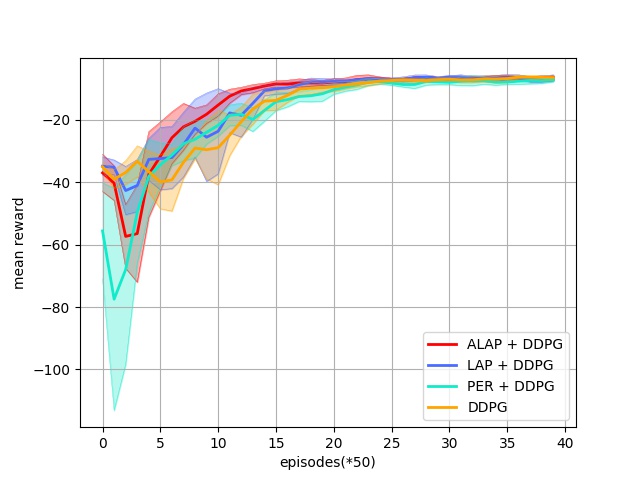}}
	\subfigure[batchsize=64]{
		\includegraphics[scale=0.34]{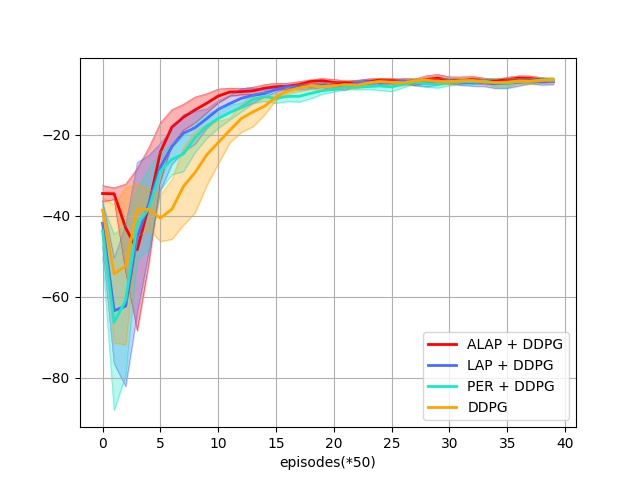}}
	\subfigure[batchsize=128]{
		\includegraphics[scale=0.34]{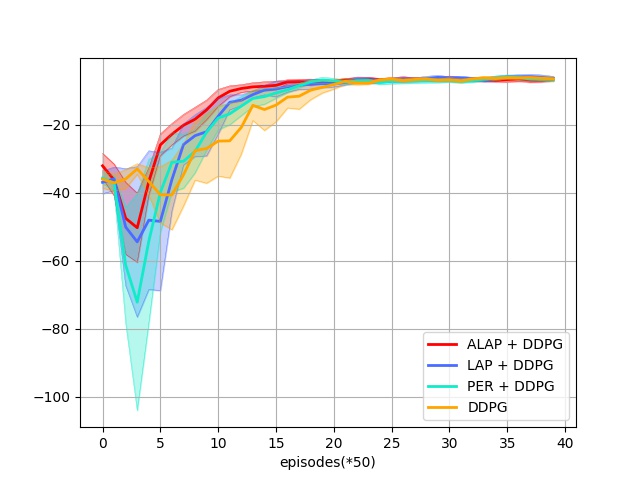}}
	\caption{Mean reward of policy gradient based algorithm.}
	\label{fig_4}
\end{figure*}

Fig.4 shows the average reward curves of ALAP + DDPG, LAP + DDPG, PER + DDPG, and the conventional DDPG. It demonstrates that ALAP has the best performance with respect to convergence speed and training variance in the face of different batchsize configurations. PER performs poorly under A-C framework, converging slower than baseline when the mini-batch is too large or too small, and PER consistently has the largest variance throughout the training process that shows the extremely instability. Compared with PER, the training variance of LAP is alleviated, but its convergence speed is not significantly improved. Overall, ALAP solves the problem of the degraded performance of PER when embedded in the A-C framework.

\subsection{Results with MADDPG}\label{MADDPG}

\iffalse
Experience Replay gives the samples in the experience pool the same sampled frequency, when the mechanism is applied to the multi-agent environment, the system state space increases exponentially with the increase of the agent's number, this uniform sampling method is bound to cause too much invalid exploration in the early stage of training. In order to solve this problem,\fi

The classical environment for Multi-to-Multi pursuit evasion game \textbf{simple$\_$tag} is used to test all the algorithms. In \textbf{simple$\_$tag}, both the pursuers and the evaders empoly a zero-sum reward structure, where the reward of the pursuers is inversely proportional to their relative distance from the evaders, while the evaders are rewarded in the opposite way. To visually compare the algorithms' strengths and weaknesses; we use MADDPG as a baseline to train evaders, while we train the pursuers with the improved algorithm simultaneously, and record their average reward. During training, we trace the model structure and hyperparameter configurations of \textbf{simple}, each agent follows the parameterized policy $\pi_i$, and 20 comparative tests were completed under each batchsize.

\iffalse
\begin{figure*}[htbp] %??
	%\rotatebox{90}{\scriptsize{~~~~~~~~~~~~~~~~~~~~~~~reward}}
	\begin{minipage}[t]{0.33\linewidth} %??????????
		\centering
		\includegraphics[width=2.5in, height=1.9in]{NEWPLOT/simpletag32.jpeg} %????????
		%\caption{Mean reward of Multi agent reinforcement learning algorithm} %?????
		\label{MADDPG} %????
	\end{minipage}%
	\begin{minipage}[t]{0.33\linewidth}
		\centering
		\includegraphics[width=2.5in, height=1.9in]{NEWPLOT/simpletag64.jpeg}
		%\caption{Simple$\_$tag}
		\label{fig2}
	\end{minipage}%
	\begin{minipage}[t]{0.33\linewidth}
		\centering
		\includegraphics[width=2.5in, height=1.9in]{NEWPLOT/simpletag128.jpeg}
		%\caption{simple$\_$adversary}
		\label{fig3}
	\end{minipage}
	\caption{Mean reward of Multi agent reinforcement learning algorithm} 
\end{figure*}
\fi

\begin{figure*}[!t]
	\centering
	\subfigure[batchsize=32]{
		\includegraphics[scale=0.34]{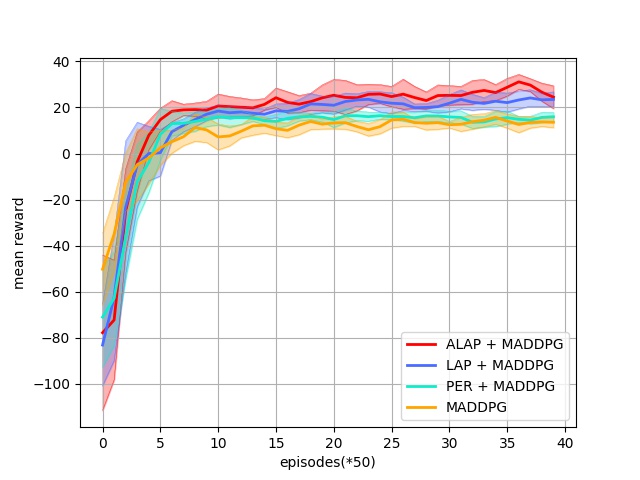}}
	\subfigure[batchsize=64]{
		\includegraphics[scale=0.34]{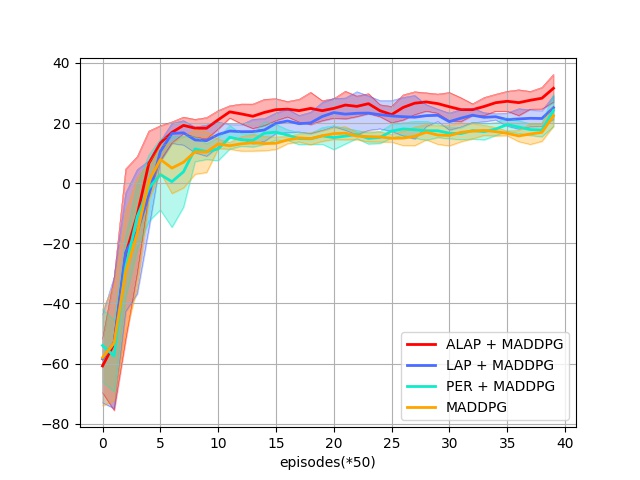}}
	\subfigure[batchsize=128]{
		\includegraphics[scale=0.34]{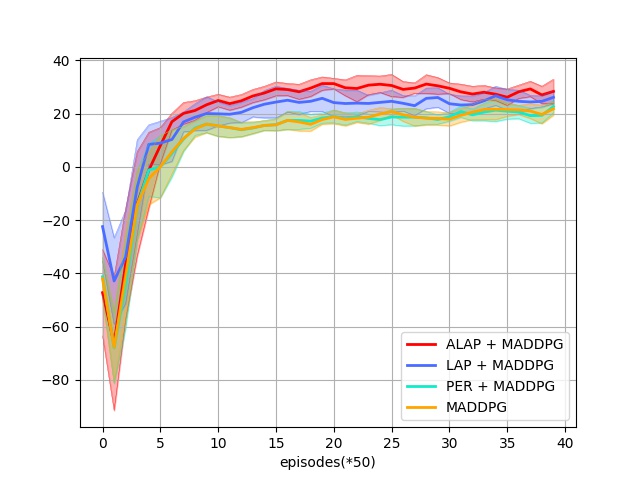}}
	\caption{Mean reward of Multi agent reinforcement learning algorithm.}
	\label{fig_5}
\end{figure*}

Fig.5 shows the average reward curves of  ALAP + MADDPG, LAP + MADDPG, PER + MADDPG, and MADDPG under different batchsize configurations. From Fig.5, we can see that under any mini-batch condition, the convergence speed and the average reward of ALAP are much higher than those of other algorithms. Although the stability of LAP decreases after mini-bach $\geq$ 64, its average reward and convergence speed are still higher than those of PER and MADDPG. The performance of PER is even worse than that of baseline after mini-batch $\geq$ 64, and the huge variance in the early stage of training is the reason that restricts the quality of its training.

\section{CONCLUSIONS}\label{conclusion}

In this paper, a general reinforcement learning algorithm framework called ALAP  is proposed, which significantly reduces the estimation deviation of Q-value function caused by PER class algorithms. Firstly, the loss function is described in segments by Huber equation, which suppresses the sensitivity of the algorithm to outliers. Furthermore, the specific progress of training is quantified by calculating the similarity of samples in experience pool through the improved Self-Attention mechanism, so as to fit the accurate value of the hyperparameter $\beta$ to regulates the importance sampling weight. In addition, we also design a Double-Sampling mechanism based on mirror buffer, and use two sampling methods simultaneously to provide data sources for network model and attention module to ensure algorithm's stable operation. The comparison results verify that ALAP can greatly improve the speed and stability of training.

\bibliographystyle{IEEEtran}

\end{document}